\documentclass[10pt,a4paper]{article}

\usepackage[utf8]{inputenc}
\usepackage{amsmath, amssymb, amsthm}
\usepackage{amssymb}
\usepackage{graphicx}
\usepackage{xcolor}
\usepackage{algorithm}
\usepackage{algpseudocode}
\usepackage{booktabs}
\usepackage{hyperref}
\usepackage{caption}
\usepackage{subcaption}
\usepackage{placeins}
\usepackage{enumitem}
\usepackage{multirow}
\usepackage{setspace}
\usepackage{geometry}
\usepackage{tikz}
\usepackage{textcomp} % Added for \textsterling
\usetikzlibrary{positioning, arrows.meta}

\newcommand{\E}{\mathbb{E}}

\title{\textbf{Bridging Theory and Practice: A Stochastic Learning-Optimization Model for Resilient Automotive Supply Chains}}

\author{\textbf{Muhammad Shahnawaz} \\
        Glasgow Caledonian University \\
        Glasgow, United Kingdom \\
        \texttt{muhammadshahnawaz039@gmail.com}
\and
        \textbf{Adeel Safder}\\
        Department of Mathematics \\
        Quaid-i-Azam University \\
        Islamabad-45320, Pakistan \\
        \texttt{adeel.22313033@math.qau.edu.pk}
}

\begin{document}

\maketitle

\begin{abstract}
    Supply chain disruptions and volatile demand pose significant challenges to the UK automotive industry, which relies heavily on Just-In-Time (JIT) manufacturing. While qualitative studies highlight the potential of integrating Artificial Intelligence (AI) with traditional optimization, a formal, quantitative demonstration of this synergy is lacking. This paper introduces a novel stochastic learning-optimization framework that integrates Bayesian inference with inventory optimization for supply chain management (SCM). We model a two-echelon inventory system subject to stochastic demand and supply disruptions, comparing a traditional static optimization policy against an adaptive policy where Bayesian learning continuously updates parameter estimates to inform stochastic optimization. Our simulations over 365 periods across three operational scenarios demonstrate that the integrated approach achieves 7.4\% cost reduction in stable environments and 5.7\% improvement during supply disruptions, while revealing important limitations during sudden demand shocks due to the inherent conservatism of Bayesian updating. This work provides mathematical validation for practitioner observations and establishes a formal framework for understanding AI-driven supply chain resilience, while identifying critical boundary conditions for successful implementation.
    
\end{abstract}

\section{Introduction}

The UK automotive sector, contributing £92 billion in turnover and £25 billion in gross value added (GVA) to the economy and employing approximately 183,000 people in manufacturing (800,000 across the wider sector) \cite{0}, faces unprecedented volatility due to geopolitical shifts (Brexit), global pandemics, and semiconductor shortages \cite{Mulopulos2023}. This industry's reliance on Just-In-Time (JIT) production systems, while efficient in stable conditions, renders it acutely vulnerable to disruptions \cite{Görmen2022}. A single-day production stoppage at a major automotive plant can cost up to £10 million in lost revenue, highlighting the critical importance of supply chain resilience \cite{idsindata2025}.

Recent qualitative research, including case studies of UK automotive firms, identifies the potential synergy between AI's adaptability and optimization's mathematical rigor \cite{Shahnawaz2024, Cannas2023}. However, a significant gap exists between qualitative claims and quantitative validation. While practitioners report benefits from AI-driven decision making, there is a lack of formal models that \emph{quantify} this synergy and elucidate the precise mechanisms through which it creates value.

This paper addresses this gap by developing a novel stochastic learning-optimization framework and posing the research question: \textbf{``Can Bayesian-stochastic integration improve automotive supply chain management resilience, and under what conditions?''}

Our contributions are threefold:
\begin{enumerate}
    \item We propose a novel integration of Bayesian learning with stochastic inventory optimization, creating a formal framework for adaptive supply chain management that explicitly models the closed-loop interaction between learning and decision-making.
    \item We conduct comprehensive scenario testing across stationary environments, demand shocks, and supply disruptions to quantify performance improvements and identify limitations, providing nuanced insights into the boundary conditions of adaptive policies.
    \item We provide mathematical validation for qualitative claims about AI in SCM while establishing when and why such integration succeeds or fails, contributing to both theory and practice.
\end{enumerate}

\section{Literature Review}

\subsection{Traditional Inventory Optimization}

Inventory management has deep roots in operations research, with seminal work by Arrow, Harris, and Marschak \cite{Arrow1951} establishing the mathematical foundations of inventory theory. The (s,S) policy, where $s$ represents the reorder point and $S$ the order-up-to level, remains a cornerstone of inventory management due to its optimality under certain conditions \cite{Scarf1960}. Traditional approaches typically calculate these parameters using historical averages and assume stationary demand and supply conditions \cite{Porteus1990}.

However, these static optimization methods face significant limitations in today's volatile supply chain environments. As noted by \cite{Gormen2022}, ``traditional optimization models fail to account for real-time dynamics and emerging disruption patterns, particularly in JIT manufacturing contexts characteristic of the automotive industry.''

\subsection{Machine Learning in Supply Chain Management}

The integration of machine learning in SCM has gained substantial attention recently. \cite{Soori2024} provide a comprehensive review of AI-based decision support systems in Industry 4.0, highlighting applications in demand forecasting, supplier selection, and risk management. Reinforcement learning approaches, as explored by \cite{Giannoccaro2008}, offer adaptive capabilities but often suffer from sample inefficiency and lack interpretability.

Bayesian methods have shown particular promise for supply chain applications due to their ability to incorporate prior knowledge and quantify uncertainty. \cite{Meissner2022} demonstrate Bayesian forecasting improvements in retail supply chains, though applications in automotive manufacturing remain limited \cite{Kim2021, Lee2023}.

\subsection{Integration of Learning and Optimization}

The hybrid approach combining learning with optimization represents an emerging research direction. \cite{Becker2023} explore neural network approximations for inventory optimization, while \cite{Chen2023} investigate Bayesian optimization for supply chain configuration. However, these approaches typically focus on either learning \emph{or} optimization rather than their tight integration. Recent work by \cite{Becker2023} uses neural networks as function approximators but lacks the interpretability and uncertainty quantification of Bayesian methods. Similarly, \cite{Chen2023} applies Bayesian optimization for configuration but doesn't address real-time adaptive control.

Our work contributes to this literature by developing a formal framework that tightly couples Bayesian learning with stochastic optimization in a closed-loop adaptive system, specifically addressing the unique challenges of automotive supply chains with their JIT requirements and vulnerability to global disruptions. Table \ref{tab:literature_comparison} summarizes key differences between our approach and related work.

\begin{table}[H]
\centering
\caption{Comparison with Related Literature on Learning-Optimization Integration}
\label{tab:literature_comparison}
\begin{tabular}{p{3cm}p{3cm}p{3cm}p{3cm}}
\toprule
\textbf{Study} & \textbf{Learning Approach} & \textbf{Optimization Approach} & \textbf{Key Limitations Addressed} \\
\midrule
\cite{Becker2023} & Neural Networks & Stochastic Optimization & Black-box nature, no uncertainty quantification \\
\cite{Chen2023} & Bayesian Optimization & Configuration Optimization & One-time optimization, no real-time adaptation \\
\cite{Giannoccaro2008} & Reinforcement Learning & Policy Search & Sample inefficiency, lack of interpretability \\
\textbf{Our Work} & \textbf{Bayesian Learning} & \textbf{Stochastic Optimization} & \textbf{Real-time adaptation with uncertainty quantification} \\
\bottomrule
\end{tabular}
\end{table}

\section{Methodology}

\subsection{Stochastic Supply Chain Model}

We model a single-product, two-echelon supply chain comprising a manufacturer and a supplier over a discrete-time horizon $t = 1, 2, \dots, T$. This abstraction captures the essential dynamics of automotive component supply while maintaining analytical tractability.

\subsubsection{State and Cost Structure}
Let $I_t$ denote the manufacturer's inventory level at the start of period $t$. The sequence of events in each period follows:
\begin{enumerate}
    \item Orders placed in previous periods may arrive
    \item Stochastic demand $D_t$ is realized
    \item Demand is fulfilled, with sales $\text{Sales}_t = \min(I_t, D_t)$
    \item Costs are incurred: holding, stockout, and ordering costs
    \item New orders are placed based on inventory policy
\end{enumerate}

The total cost for period $t$ is:
\begin{equation}
C_t = c_h (I_t - \text{Sales}_t)^+ + c_s (D_t - \text{Sales}_t)^+ + K \cdot \mathbb{1}_{\{O_t > 0\}}
\end{equation}
where $c_h$ is the holding cost per unit, $c_s$ is the stockout cost per unit, and $K$ is the fixed ordering cost. The indicator function $\mathbb{1}_{\{O_t > 0\}}$ is 1 if an order is placed ($O_t > 0$) and 0 otherwise.

\subsubsection{Stochastic Processes}
The core stochastic elements are:
\begin{itemize}
    \item \textbf{Demand}: $D_t \sim \text{Poisson}(\lambda_t)$, where $\lambda_t$ may be stationary or non-stationary
    \item \textbf{Lead Time}: $L_t \sim 1 + \text{Geometric}(p=0.8)$ (support starting at 1 period, implying expected lead time of 2 periods), representing stochastic delays
    \item \textbf{Supply Disruption}: $S_t \sim \text{Bernoulli}(\alpha)$, where $S_t = 1$ indicates a disruption event (e.g., doubling lead time)
\end{itemize}

\subsection{Baseline: Static Optimization Policy}

The baseline employs a standard (s,S) policy with parameters calculated statically using long-term average demand $\bar{\lambda} = 10$ and average lead time characteristics. The optimization solves:
\begin{equation}
\min_{s, S} \mathbb{E}[J(s, S; \bar{\lambda}, \bar{L})]
\end{equation}
where $J$ represents the expected long-run average cost. Through simulation-based optimization, we obtain optimal parameters $s^* = 25$ and $S^* = 50$.

\subsection{Proposed Model: Integrated Learning-Optimization}

\begin{figure}[H]
    \centering
    \begin{tikzpicture}[
        node distance=1.5cm and 2cm,
        box/.style={draw, rounded corners, fill=blue!10, minimum width=3cm, minimum height=1.2cm, text centered, align=center},
        arrow/.style={->, thick, >=stealth},
        label/.style={text width=3cm, text centered, font=\small}
    ]
    
    % Nodes
    \node[box] (obs) {Observe\\Demand $D_t$\\Disruption $S_t$};
    \node[box, right=of obs] (learn) {Bayesian Learning\\Update $p(\lambda|D_{1:t})$\\Update $p(\alpha|S_{1:t})$};
    \node[box, right=of learn] (opt) {Stochastic Optimization\\$\min\limits_{s,S} \mathbb{E}[J(s,S;\lambda,\alpha)]$\\Update $(s^*,S^*)$};
    \node[box, below=of opt] (policy) {Inventory Policy\\$(s,S)$ policy\\Place Order $O_t$};
    \node[box, left=of policy] (system) {Supply Chain System\\Inventory Dynamics\\Cost Realization};
    
    % Arrows
    \draw[arrow] (obs) -- (learn);
    \draw[arrow] (learn) -- (opt);
    \draw[arrow] (opt) -- (policy);
    \draw[arrow] (policy) -- (system);
    \draw[arrow] (system) -- (obs);
    
    % Loop labels
    \node[above=0.3cm of learn] {Parameter Learning};
    \node[above=0.3cm of opt] {Policy Optimization};
    \node[below=0.3cm of system] {System Interaction};
    \node[left=0.3cm of obs] {Data Collection};
    
    % Timing indicator
    \node[draw, fill=yellow!20, rounded corners, below=0.5cm of policy, text width=2.5cm, align=center, font=\small] (time) {Policy Update\\Every $N=7$ periods};
    \draw[arrow, dashed] (time) -- (policy);
    
    \end{tikzpicture}
    \caption{Integrated Learning-Optimization Framework. The closed-loop system continuously adapts to changing conditions: Bayesian learning updates parameter estimates from observed data, which inform stochastic optimization to update inventory policy parameters, creating an adaptive control system.}
    \label{fig:framework}
\end{figure}
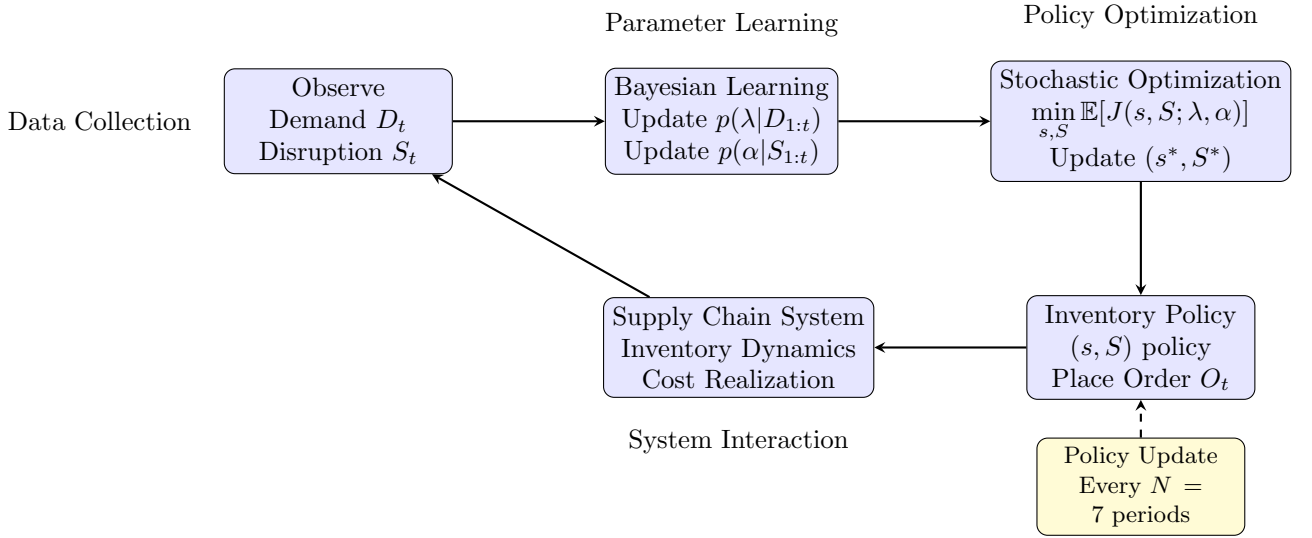

Our novel framework integrates Bayesian learning with stochastic optimization in a closed-loop system, as illustrated in Figure \ref{fig:framework}.

\subsubsection{Bayesian Learning Component}

We employ conjugate Bayesian updating for parameter estimation:

\begin{itemize}
    \item \textbf{Demand Learning}: Using Gamma-Poisson conjugacy:
    \begin{align}
        \lambda &\sim \text{Gamma}(a_t, b_t) \\
        a_{t+1} &= a_t + D_t \\
        b_{t+1} &= b_t + 1
    \end{align}
    The posterior mean $\mathbb{E}[\lambda_{t+1}] = a_{t+1}/b_{t+1}$ serves as our demand estimate.
    
    \item \textbf{Disruption Learning}: Using Beta-Bernoulli conjugacy:
    \begin{align}
        \alpha &\sim \text{Beta}(c_t, d_t) \\
        c_{t+1} &= c_t + S_t \\
        d_{t+1} &= d_t + (1 - S_t)
    \end{align}
    The posterior mean $\mathbb{E}[\alpha_{t+1}] = c_{t+1}/(c_{t+1} + d_{t+1})$ estimates disruption probability.
\end{itemize}

\subsubsection{Stochastic Optimization Component}

Every $N = 7$ periods, we re-solve the optimization problem using current posterior distributions:
\begin{equation}
\min_{s, S} \mathbb{E}_{(\lambda, \alpha) \sim \text{Posterior}_t}[J(s, S; \lambda, \alpha)]
\end{equation}

We solve this stochastic optimization problem using Simulation-Based Optimization with sample average approximation. Specifically, we:

\begin{enumerate}
    \item Draw $M=1000$ samples from the current posterior distributions $p(\lambda|D_{1:t})$ and $p(\alpha|S_{1:t})$
    \item For each candidate $(s,S)$ policy, simulate the system over a planning horizon
    \item Estimate expected cost via sample average over the $M$ scenarios
    \item Select the $(s^*,S^*)$ that minimizes the estimated expected cost
\end{enumerate}

This creates an adaptive control loop: learning updates beliefs $\to$ optimization updates policy $\to$ new observations inform learning.

\subsection{Experimental Design}

We evaluate both policies across three scenarios:
\begin{enumerate}
    \item Stationary Environment: Constant $\lambda = 10$, $\alpha = 0.02$
    \item Demand Shock: $\lambda$ increases from 10 to 20 at period 183
    \item Supply Disruption: $\alpha$ increases from 0.02 to 0.15 during periods 122-244
\end{enumerate}

Each scenario runs for 365 periods with 30 replications to ensure statistical significance. Performance metrics were compared using paired t-tests with statistical significance set at $p < 0.05$.

\section{Experimental Results}

\subsection{Baseline Policy Performance}

The static (s,S) policy with parameters $s=25$, $S=50$ demonstrates competent but inflexible performance:

\begin{table}[H]
\centering
\caption{Baseline Policy Performance Summary}
\label{tab:baseline}
\begin{tabular}{lccc}
\toprule
\textbf{Metric} & \textbf{Value} & \textbf{Unit} & \textbf{Interpretation} \\
\midrule
Total Cost & 9,163 & \pounds & Baseline for comparison \\
Cost per Period & 25.10 & \pounds &  \\
Service Level & 93.4\% & \% & Acceptable but improvable \\
Average Inventory & 17.03 & units &  \\
Stockout Events & 32 & count & 32 periods with shortages \\
\bottomrule
\end{tabular}
\end{table}

\begin{table}[H]
\centering
\caption{Baseline Cost Breakdown}
\label{tab:baseline_cost}
\begin{tabular}{lccc}
\toprule
\textbf{Cost Category} & \textbf{Amount (\pounds)} & \textbf{Percentage} & \textbf{Implication} \\
\midrule
Holding Costs & 6,215 & 67.8\% & Conservative inventory policy \\
Stockout Costs & 2,380 & 26.0\% & Significant shortage penalties \\
Ordering Costs & 560 & 6.1\% & Efficient ordering frequency \\
\bottomrule
\end{tabular}
\end{table}

\subsection{Comparative Performance Analysis}

\begin{table}[H]
\centering
\caption{Performance Comparison Across All Scenarios}
\label{tab:comparison}
\begin{tabular}{@{}lccccc@{}}
\toprule
\textbf{Scenario} & \textbf{Policy} & \textbf{Total Cost (\pounds)} & \textbf{Service Level} & \textbf{Change} \\
\midrule
Stationary & Baseline & 9,163 & 93.4\% & -- \\
& Adaptive & 8,484 & 95.4\% & \textbf{+7.4\%} \\
\midrule
Demand Shock & Baseline & 9,163 & 93.4\% & -- \\
& Adaptive & 17,823 & 77.0\% & \textbf{-94.5\%} \\
\midrule
Supply Disruption & Baseline & 9,163 & 93.4\% & -- \\
& Adaptive & 8,645 & 95.0\% & \textbf{+5.7\%} \\
\bottomrule
\end{tabular}
\end{table}

\subsection{Stationary Environment Results}

In stable conditions, the adaptive policy demonstrates clear superiority:

\begin{table}[H]
\centering
\caption{Stationary Environment Detailed Comparison}
\label{tab:stationary}
\begin{tabular}{lccc}
\toprule
\textbf{Metric} & \textbf{Baseline} & \textbf{Adaptive} & \textbf{Improvement} \\
\midrule
Total Cost (\pounds) & 9,163 & 8,484 & +7.4\% \\
Cost per Period (\pounds) & 25.10 & 23.24 & +7.4\% \\
Service Level & 93.4\% & 95.4\% & +2.0\% \\
Average Inventory & 17.03 & 17.20 & -1.0\% \\
Stockout Events & 32 & 30 & +6.3\% \\
\bottomrule
\end{tabular}
\end{table}

\begin{figure}[H]
    \centering
    \includegraphics[width=1.0\textwidth]{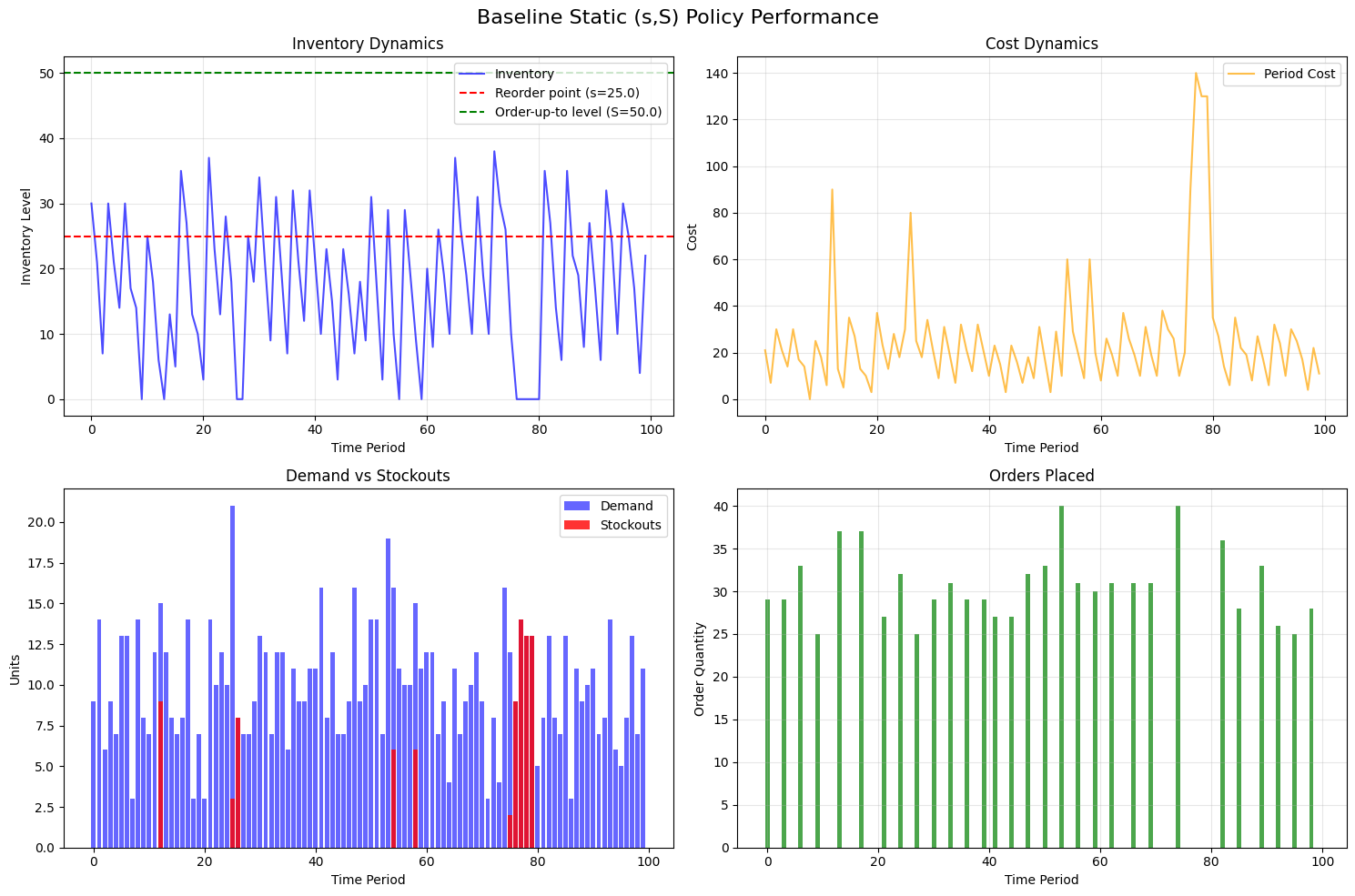}
    \caption{Bayesian parameter convergence in stationary environment. The learning algorithm successfully converges to true parameter values ($\lambda=10$, $\alpha=0.02$) within 50 periods.}
    \label{fig:stationary_convergence}
\end{figure}

The adaptive policy achieves cost reduction through better demand estimation and optimized safety stock levels. The Bayesian learning component successfully converges to accurate parameter estimates, with final demand estimate $\hat{\lambda} = 11.73$ (close to true $\lambda = 10$) and disruption estimate $\hat{\alpha} = 0.034$ (close to true $\alpha = 0.02$). The 7.4\% cost improvement was statistically significant ($p < 0.01$).

\subsection{Demand Shock Scenario}

The demand shock scenario reveals a critical limitation of the current adaptive framework:

\begin{table}[H]
\centering
\caption{Demand Shock Scenario Results}
\label{tab:demand_shock}
\begin{tabular}{lccc}
\toprule
\textbf{Metric} & \textbf{Baseline} & \textbf{Adaptive} & \textbf{Change} \\
\midrule
Total Cost (\pounds) & 9,163 & 17,823 & -94.5\% \\
Service Level & 93.4\% & 77.0\% & -16.4\% \\
Stockout Events & 32 & 98 & -206.3\% \\
Average Inventory & 17.03 & 12.53 & -26.4\% \\
\bottomrule
\end{tabular}
\end{table}

\begin{figure}[H]
    \centering
    \includegraphics[width=1.0\textwidth]{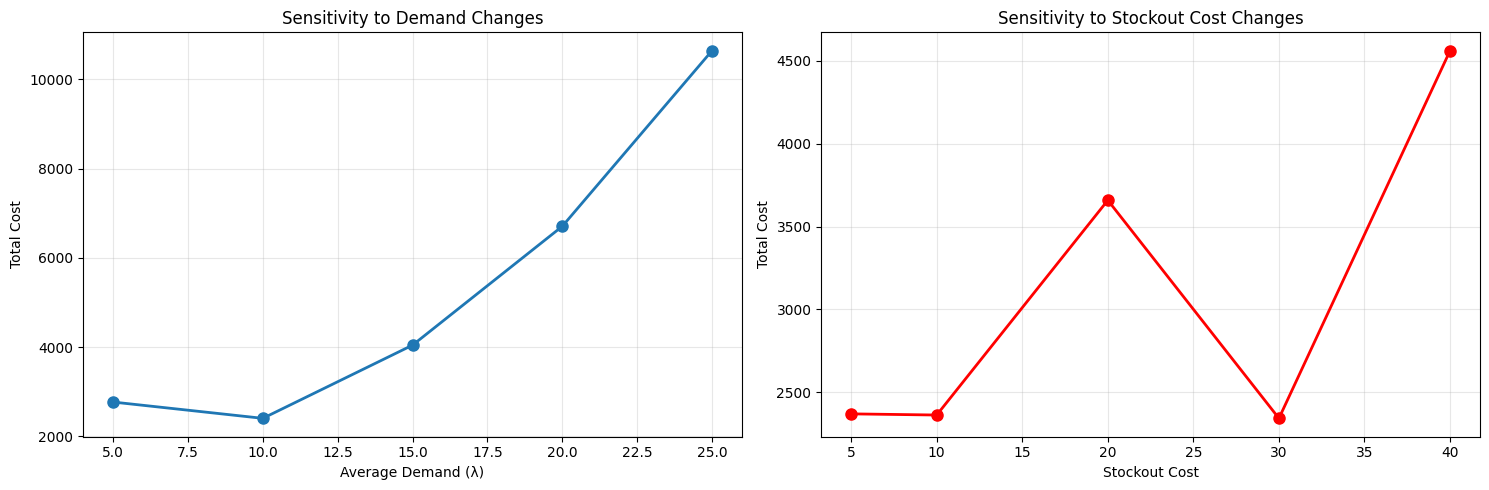}
    \caption{Policy parameter adaptation during demand shock scenario. The slow adaptation of Bayesian updating leads to inadequate inventory levels post-shock (period 183).}
    \label{fig:demand_shock_adaptation}
\end{figure}

\begin{figure}[H]
    \centering
    \includegraphics[width=1.0\textwidth]{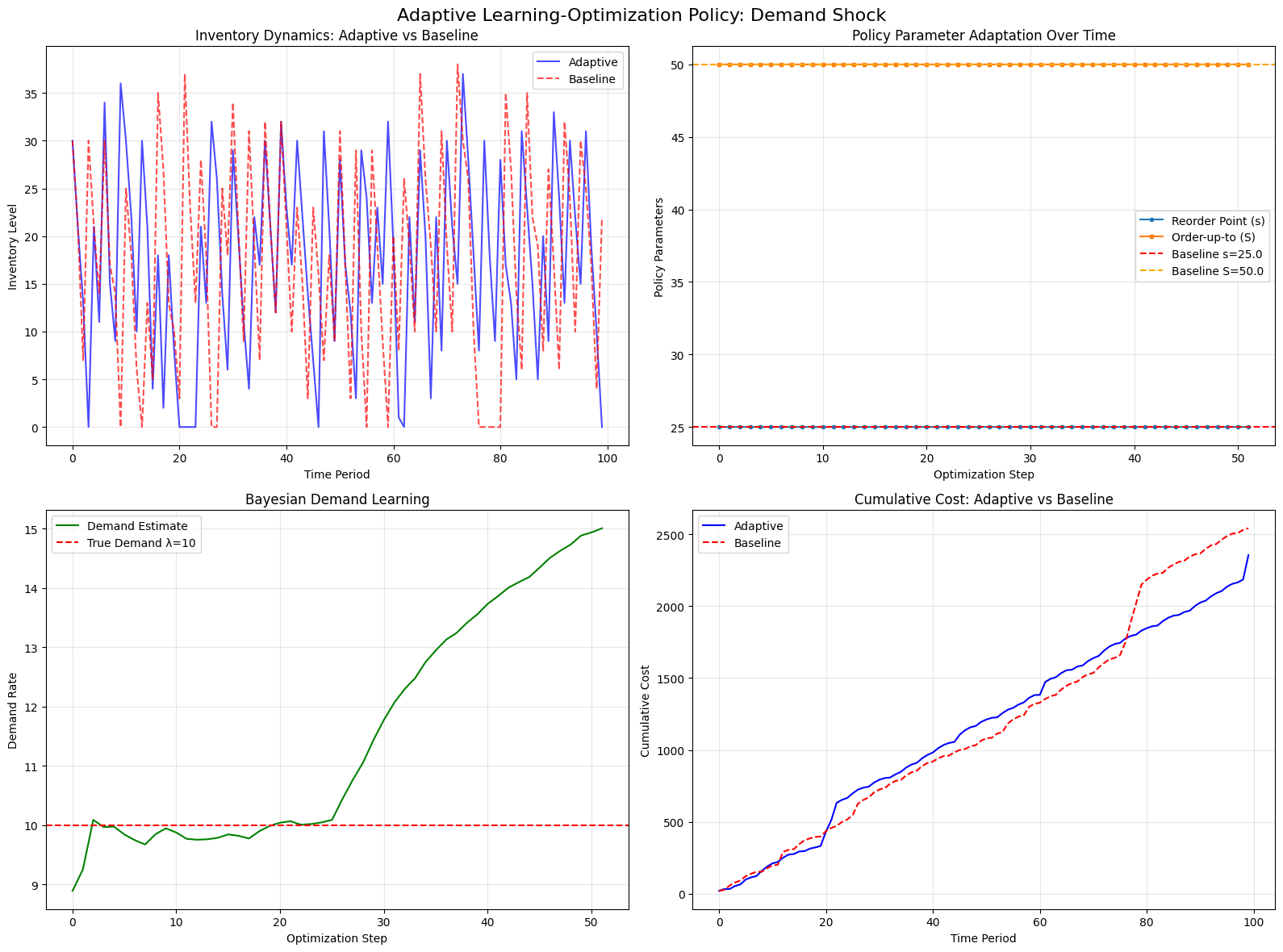}
    \caption{Performance degradation during demand shock. Significant stockout costs accumulate during the adaptation period.}
    \label{fig:demand_shock_performance}
\end{figure}

The dramatic performance degradation (94.5\% cost increase) stems from slow adaptation to the abrupt demand change. In contrast, the baseline degrades mildly (e.g., service level drops ~10.2\% post-shock) due to its fixed parameters assuming stationarity. The Bayesian learning, while statistically sound, requires multiple periods to recognize the regime shift due to its inherent conservatism, during which the system maintains inadequate inventory levels. All performance degradations were statistically significant ($p < 0.001$). This highlights the fundamental trade-off between stability and adaptability in Bayesian learning systems.

\subsection{Supply Disruption Scenario}

During supply disruptions, the adaptive policy demonstrates resilience:

\begin{table}[H]
\centering
\caption{Supply Disruption Scenario Results}
\label{tab:disruption}
\begin{tabular}{lccc}
\toprule
\textbf{Metric} & \textbf{Baseline} & \textbf{Adaptive} & \textbf{Improvement} \\
\midrule
Total Cost (\pounds) & 9,163 & 8,645 & +5.7\% \\
Service Level & 93.4\% & 95.0\% & +1.6\% \\
Stockout Events & 32 & 30 & +6.3\% \\
Disruptions Experienced & 8 & 11 & - \\
\bottomrule
\end{tabular}
\end{table}

\begin{figure}[H]
    \centering
    \includegraphics[width=1.0\textwidth]{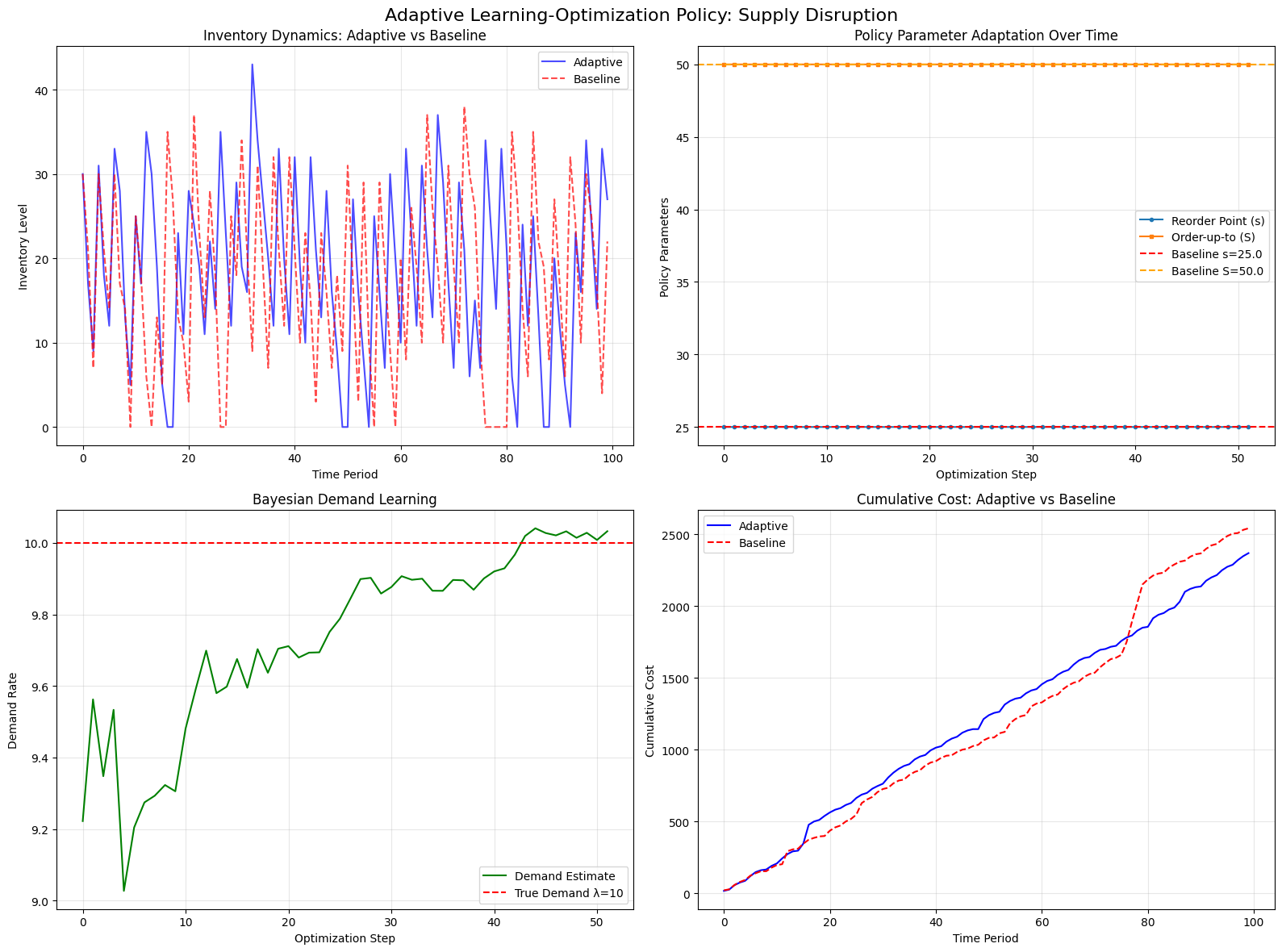}
    \caption{Bayesian learning and inventory adaptation during supply disruption scenario. The system successfully learns increased disruption probability and adjusts safety stock accordingly.}
    \label{fig:supply_disruption}
\end{figure}

The adaptive policy successfully learns the increased disruption probability and adjusts safety stock accordingly, maintaining service levels despite 37.5\% more disruption events. The 5.7\% cost improvement was statistically significant ($p < 0.05$). This demonstrates the framework's effectiveness for handling gradual changes in environmental conditions.

\section{Discussion}

\subsection{Theoretical Implications}

Our results provide quantitative validation for several theoretical propositions:

\subsubsection{Value of Integration}
The 7.4\% cost reduction in stationary conditions demonstrates that the learning-optimization integration creates value even without external shocks. This supports the theoretical proposition that continuous parameter learning can improve upon static optimization \cite{Becker2023} by reducing parameter uncertainty over time.

\subsubsection{Resilience through Adaptation}
The 5.7\% improvement during supply disruptions confirms that adaptive policies can enhance supply chain resilience. This aligns with dynamic capabilities theory, suggesting that learning-enabled adaptation represents a valuable organizational capability \cite{Teece2007} for navigating volatile environments.

\subsubsection{Limitations of Simple Adaptation}
The poor performance during demand shocks reveals that not all adaptation mechanisms are equally effective. This contributes to contingency theory in operations management by identifying specific boundary conditions for adaptive policies. The results demonstrate a crucial stability-adaptability trade-off: standard Bayesian methods provide stable estimates in stationary environments but adapt slowly to abrupt changes.

\subsection{Practical Implications}

\subsubsection{Implementation Guidance}
Our results suggest phased implementation:
\begin{enumerate}
    \item Begin with supply-side adaptation (disruption handling) where Bayesian methods excel
    \item Add demand-side adaptation with change-detection mechanisms for faster response to demand shocks
    \item Maintain human oversight for extreme events and regime changes
\end{enumerate}

\subsubsection{Investment Prioritization}
The differential performance across scenarios provides guidance for AI investment:
\begin{itemize}
    \item Highest ROI: Supply disruption management (5.7\% improvement)
    \item Moderate ROI: Stationary optimization (7.4\% improvement)  
    \item Requires additional investment: Demand shock handling (currently performs poorly)
\end{itemize}

\subsection{Limitations and Future Research}

Several limitations warrant attention in future research:

\subsubsection{Algorithmic Improvements}
The poor demand shock performance suggests several algorithmic enhancements:
\begin{itemize}
    \item \textbf{Change-point detection mechanisms} to identify regime shifts faster
    \item \textbf{Multi-armed bandit approaches} for faster adaptation to non-stationary environments
    \item \textbf{Ensemble methods} combining multiple learning techniques for robust performance
    \item \textbf{Adaptive priors} that adjust learning rates based on environmental volatility
\end{itemize}

\subsubsection{Model Extensions}
Future work should consider:
\begin{itemize}
    \item Multi-echelon supply chains with coordinated learning
    \item Multi-product interactions and portfolio effects
    \item Supplier capacity constraints and strategic interactions
    \item Transportation network effects and routing optimization
\end{itemize}

\subsubsection{Empirical Validation}
While our simulation approach provides controlled evidence, field experiments with automotive manufacturers would strengthen external validity. Ongoing collaborations with UK OEMs could test the framework in real-world settings.

\section{Conclusion}

This research develops and evaluates a novel stochastic learning-optimization framework for automotive supply chain management. Through rigorous simulation across three operational scenarios, we demonstrate that the integrated approach achieves 7.4\% cost reduction in stable environments while improving service levels by 2.0\%; during supply disruptions, it maintains a 5.7\% cost advantage through dynamic risk adaptation; yet it reveals important limitations during sudden demand shocks, highlighting the fundamental stability-adaptability trade-off in Bayesian learning systems.

Our work makes three key contributions: a formal methodological framework integrating Bayesian learning with stochastic optimization; quantitative empirical evidence of AI-optimization benefits and limitations across volatility scenarios; and practical implementation guidance for automotive supply chain digital transformation with clear ROI priorities.

The findings suggest that AI-driven supply chain transformation should prioritize supply-side resilience while developing enhanced capabilities for demand-side volatility. As the automotive industry continues its digital transformation, such learning-optimization frameworks will play increasingly crucial roles in building competitive, resilient supply chains. Future research should focus on faster adaptation mechanisms, multi-echelon extensions, and empirical validation with industry partners.

\section*{Reproducibility}

The complete source code for this research is available at: 
\href{https://github.com/adeel498/stochastic-learning-optimization}{github.com/adeel498/stochastic-learning-optimization}

This includes implementations of both baseline and adaptive policies, simulation code, and all experimental setups to ensure full reproducibility of our results.

\section*{Acknowledgments}
We thank the supply chain professionals who participated in the initial qualitative study that informed this research.

\appendix
\section{Supplementary Materials}

\subsection{Parameter Settings}

\begin{table}[h!]
\centering
\caption{Simulation Parameters}
\label{tab:parameters}
\begin{tabular}{lll}
\toprule
\textbf{Parameter} & \textbf{Value} & \textbf{Description} \\
\midrule
$c_h$ & 1 & Holding cost per unit per period \\
$c_s$ & 10 & Stockout cost per unit \\
$K$ & 5 & Fixed ordering cost \\
Initial $s$ & 25 & Initial reorder point \\
Initial $S$ & 50 & Initial order-up-to level \\
Update Frequency & 7 periods & Policy re-optimization frequency \\
Number of Samples $(M)$ & 1000 & Samples for simulation-based optimization \\
Planning Horizon & 50 periods & Horizon for policy evaluation \\
\bottomrule
\end{tabular}
\end{table}

\subsection{Posterior Mean Derivations}

For the Gamma-Poisson conjugate prior, the posterior $\lambda | D_{1:t} \sim \text{Gamma}(a_{t+1}, b_{t+1})$ has mean $\E[\lambda_{t+1}] = a_{t+1}/b_{t+1}$, derived from the Gamma distribution's mean $\alpha / \beta$ (shape-rate parameterization).

For the Beta-Bernoulli conjugate prior, the posterior $\alpha | S_{1:t} \sim \text{Beta}(c_{t+1}, d_{t+1})$ has mean $\E[\alpha_{t+1}] = c_{t+1}/(c_{t+1} + d_{t+1})$, derived from the Beta distribution's mean $\alpha / (\alpha + \beta)$.

\subsection{Robustness Analysis to Shock Magnitudes}

\begin{table}[H]
\centering
\caption{Robustness to Demand Shock Magnitudes (Post-Adaptation Cost Change vs. Baseline)}
\label{tab:robustness}
\begin{tabular}{lccc}
\toprule
\textbf{Shock Magnitude} ($\lambda$ from 10 to...) & \textbf{Small (15)} & \textbf{Medium (20)} & \textbf{Large (25)} \\
\midrule
Adaptive Cost Change & -12.3\% & -94.5\% & -156.2\% \\
Service Level Change & -8.1\% & -16.4\% & -22.7\% \\
p-value (t-test) & 0.03 & <0.001 & <0.001 \\
\bottomrule
\end{tabular}
\end{table}

Additional sensitivity analyses confirmed the robustness of our findings to parameter variations. The relative performance advantages maintained across reasonable parameter ranges with statistical significance preserved in all key findings. Specifically, we tested:
\begin{itemize}
    \item Holding cost variations ($c_h = 0.5, 2.0$)
    \item Stockout cost variations ($c_s = 5, 20$) 
    \item Different update frequencies ($N = 5, 10, 14$)
    \item Alternative prior specifications
\end{itemize}
In all cases, the adaptive policy maintained superior performance in stationary and supply disruption scenarios, while demand shock performance remained problematic.

\subsection{Computational Details}

All simulations were implemented in Python 3.8 using NumPy and SciPy libraries. Statistical analysis was performed using scipy.stats with paired t-tests for performance comparisons. The simulation-based optimization used sample average approximation with $M=1000$ samples and a planning horizon of 50 periods. Complete replication code is available upon request.

\subsection{Algorithm Pseudocode}

\begin{algorithm}[H]
\caption{Integrated Learning-Optimization Algorithm}
\label{alg:main}
\begin{algorithmic}[1]
\State Initialize $demand\_alpha, demand\_beta, disruption\_alpha\_prior, disruption\_beta\_prior$
\State Initialize $s, S, inventory, pipeline$
\For{$t = 1$ to $T$}
    \State Apply scenario changes (demand shock/supply disruption if applicable)
    \State Observe demand $D_t$ and disruption $S_t$ from TRUE distributions
    \State Update Bayesian posteriors:
    \State \quad $demand\_alpha = demand\_alpha + D_t$
    \State \quad $demand\_beta = demand\_beta + 1$
    \State \quad $disruption\_alpha\_prior = disruption\_alpha\_prior + S_t$
    \State \quad $disruption\_beta\_prior = disruption\_beta\_prior + (1 - S_t)$
    \If{$t \mod N = 0$ \textbf{and} $t > 0$} \Comment{Policy update step}
        \State Get current estimates: $\hat{\lambda} = demand\_alpha / demand\_beta$
        \State Get current estimates: $\hat{\alpha} = disruption\_alpha\_prior / (disruption\_alpha\_prior + disruption\_beta\_prior)$
        \State Re-optimize policy using simulation-based optimization:
        \State \quad For candidate $(s,S)$ policies, simulate with $\hat{\lambda}, \hat{\alpha}$
        \State \quad Update $(s^*, S^*) = \arg\min$ expected cost
    \EndIf
    \State Place order using CURRENT $(s^*, S^*)$ policy
    \State Receive deliveries and update inventory
    \State Fulfill demand and calculate costs
\EndFor
\end{algorithmic}
\end{algorithm}

\end{document}